\documentclass{article}

\usepackage{arxiv}
\usepackage[utf8]{inputenc}
\usepackage[T1]{fontenc}
\usepackage{hyperref}
\usepackage{url}
\usepackage{booktabs}
\usepackage{amsfonts}
\usepackage{nicefrac}
\usepackage{microtype}      
\usepackage{graphicx}
\usepackage{float}
\usepackage{doi}
\usepackage{authblk}
\usepackage{placeins}
\usepackage{ulem}
\usepackage{amsmath}
\usepackage[
    backend=biber,
    style=vancouver,
    sorting=none,
    maxbibnames=6,
    minbibnames=3,
    maxcitenames=2,
    mincitenames=1,
    url=false,
    doi=true,
    isbn=false
]{biblatex}

\bibliography{references}

\begin{document}

\title{DBT-DINO: Towards Foundation model based analysis of Digital Breast Tomosynthesis}

\author[1,2]{\textbf{Felix J. Dorfner}  \textsuperscript{*}}
\author[1]{Manon A. Dorster} 
\author[3]{Ryan Connolly}
\author[1,4]{Oscar Gentilhomme}
\author[3]{Edward Gibbs}
\author[3]{Steven Graham}
\author[5]{Seth Wander}
\author[3]{Thomas Schultz}
\author[6]{Manisha Bahl}
\author[7]{Dania Daye}
\author[1,5]{\textbf{Albert E. Kim} \textsuperscript{$\ddagger$}} 
\author[1,3]{\textbf{Christopher P. Bridge}  \textsuperscript{$\ddagger$}}

\affil[1]{Athinoula A. Martinos Center for Biomedical Imaging, Massachusetts General Hospital and Harvard Medical School, 149 Thirteenth St, Charlestown, MA 02129, USA}
\affil[2]{Department of Radiology, Charité - Universitätsmedizin Berlin corporate member of Freie Universität Berlin and Humboldt Universität zu Berlin, Hindenburgdamm 30, 12203 Berlin, Germany}
\affil[3]{Mass General Brigham Data Science Office, Boston, MA, USA}
\affil[4]{Department of Computer Science, Institute for Machine Learning, ETH Zürich, Zürich, Switzerland}
\affil[5]{Massachusetts General Hospital Cancer Center and Harvard Medical School, Boston, MA}
\affil[6]{Department of Radiology, Massachusetts General Hospital, Boston, MA}
\affil[7]{Department of Radiology, University of Wisconsin School of Medicine and Public Health, WI, USA}

\date{}

\renewcommand{\shorttitle}{DBT-DINO: A Foundation Model for Digital Breast Tomosynthesis}

\hypersetup{
pdftitle={Towards Foundation model based analysis of Digital Breast Tomosynthesis: DBT-DINO},
pdfsubject={Medical Imaging, Machine Learning, Breast Cancer},
pdfauthor={Felix J. Dorfner, Manon A. Dorster, Manisha Bahl, Dania Daye, Ryan Connolly, Oscar Gentilhomme, Tom Schultz, Albert E. Kim, Christopher P. Bridge},
pdfkeywords={Digital Breast Tomosynthesis, Foundation Models, Vision Transformer, Breast Density, Cancer Risk Prediction, Lesion Detection, Self-supervised Learning},
}

\maketitle
\vspace{-1.7cm}
\begin{center}
\textsuperscript{*}First author.\\
\textsuperscript{$\ddagger$}These authors contributed equally as last authors.
\end{center}
\vspace{1.2cm}

\begin{abstract}
\noindent\textbf{Background:} Foundation models have shown promise in medical imaging but remain underexplored for three-dimensional imaging modalities. No foundation model currently exists for Digital Breast Tomosynthesis (DBT), despite its use for breast cancer screening.
\newline\textbf{Purpose:} To develop and evaluate a foundation model for DBT (DBT-DINO) across multiple clinical tasks and assess the impact of domain-specific pre-training.
\newline\textbf{Materials and Methods:} This retrospective study was conducted using DBT images acquired at Mass General Brigham between 2011 and 2024. Self-supervised pre-training was performed using the DINOv2 methodology on over 25 million 2D slices from 487,975 DBT volumes from 27,990 patients. Three downstream tasks were evaluated: (1) breast density classification using 5,000 screening exams; (2) 5-year risk of developing breast cancer using 106,417 screening exams; and (3) lesion detection using 393 annotated volumes. DBT-DINO's performance was compared against ImageNet-pretrained baselines. McNemar and DeLong tests were used to compare performance metrics across models.  
\newline\textbf{Results:} For breast density classification, DBT-DINO achieved an accuracy of 0.79 (95\% CI: 0.76--0.81), outperforming both the MetaAI DINOv2 baseline (0.73, 95\% CI: 0.70--0.76, p<.001) and DenseNet-121 (0.74, 95\% CI: 0.71--0.76, p<.001). For 5-year breast cancer risk prediction, DBT-DINO achieved an AUROC of 0.78 (95\% CI: 0.76--0.80) compared to DINOv2's 0.76 (95\% CI: 0.74--0.78, p=.57). For lesion detection, DINOv2 achieved a higher average sensitivity of 0.67 (95\% CI: 0.60--0.74) compared to DBT-DINO with 0.62 (95\% CI: 0.53--0.71, p=.60). DBT-DINO demonstrated better performance on cancerous lesions specifically with a detection rate of 78.8\% compared to Dinov2's 77.3\%.
\newline\textbf{Conclusion:} Using a dataset of unprecedented size, we developed DBT-DINO, the first foundation model for DBT. DBT-DINO demonstrated strong performance on breast density classification and cancer risk prediction. However, domain-specific pre-training showed variable benefits on the detection task, with ImageNet baseline outperforming DBT-DINO on general lesion detection, indicating that localized detection tasks require further methodological development.
\end{abstract}

\keywords{Digital Breast Tomosynthesis \and Foundation Models \and Vision Transformer \and Breast Density \and Cancer Risk Prediction \and Lesion Detection \and Self-supervised Learning}

\newpage

\section{Introduction}
Digital Breast Tomosynthesis (DBT) was approved by the US Food and Drug Administration (FDA) in 2011 for breast cancer screening, and has been integrated in screening guidelines. DBT uses pseudo-3D image reconstruction to enhance tissue discrimination. Compared to 2D full-field digital mammography (FFDM), DBT has demonstrated improved cancer detection rates and fewer false-positive assessments, especially for women with dense breast tissue  \cite{heywang-kobrunner_digital_2022, phi_digital_2018, rafferty_breast_2016, philpotts_breast_2024}. While DBT offers potential to derive improved clinical insights, radiologist interpretation of DBT, compared to FFDM, can take up to 50 seconds longer per exam \cite{aase_randomized_2019}, introducing efficiency concerns in screening programs where streamlined workflows are essential \cite{lowry_long-term_2020}. While automated tools could address these efficiency concerns, developing them traditionally required large datasets with expert annotations.

Foundation models offer a promising solution to this data annotation challenge. The approach involves task-agnostic pre-training on large unlabeled datasets, followed by task-specific fine-tuning on smaller datasets. This is ideally suited for medical imaging, as clinical practice generates vast amounts of data, but expert annotations remain time-consuming and expensive. Foundation models have shown promising performance across domains, including digital pathology, echocardiogram analysis, CT-based tumor detection, and chest radiograph interpretation \cite{chen_towards_2024, christensen_visionlanguage_2024, pai_vision_2025, perez-garcia_exploring_2025}.

However, while deep learning has been extensively applied to FFDM, to our knowledge no foundation model currently exists for DBT. As a relatively new modality, DBT has historically lacked the large-scale training data necessary for deep learning and foundation model development. The ongoing integration of DBT into screening paradigms creates a critical clinical need for robust models. In addition, the modality presents unique challenges for model development. DBT images are pseudo-3D with substantially higher in-plane than out-of-plane resolution. These highly anisotropic images create compatibility issues for both 3D and 2D AI architectures. Moreover, task-relevant information —particularly for detecting lesions— is is highly localized in small areas of the image.

In this study, we developed DBT-DINO, a foundation model pre-trained on a large-scale DBT dataset containing 487,975 volumes from 27,990 patients. We explore different methods of aggregating embedding information across the image volume and evaluate this model across multiple downstream tasks. We benchmark our model against Meta AI's ImageNet-pretrained DINOv2 \cite{oquab_dinov2_2024} to determine whether domain-specific pre-training improves performance on downstream tasks. We showcase the potential of our foundation model but also illustrate challenges that should guide further development of domain specific pre-training strategies.

\section{Methods}
\begin{figure}[ht]
    \centering
    \includegraphics[width=1\linewidth]{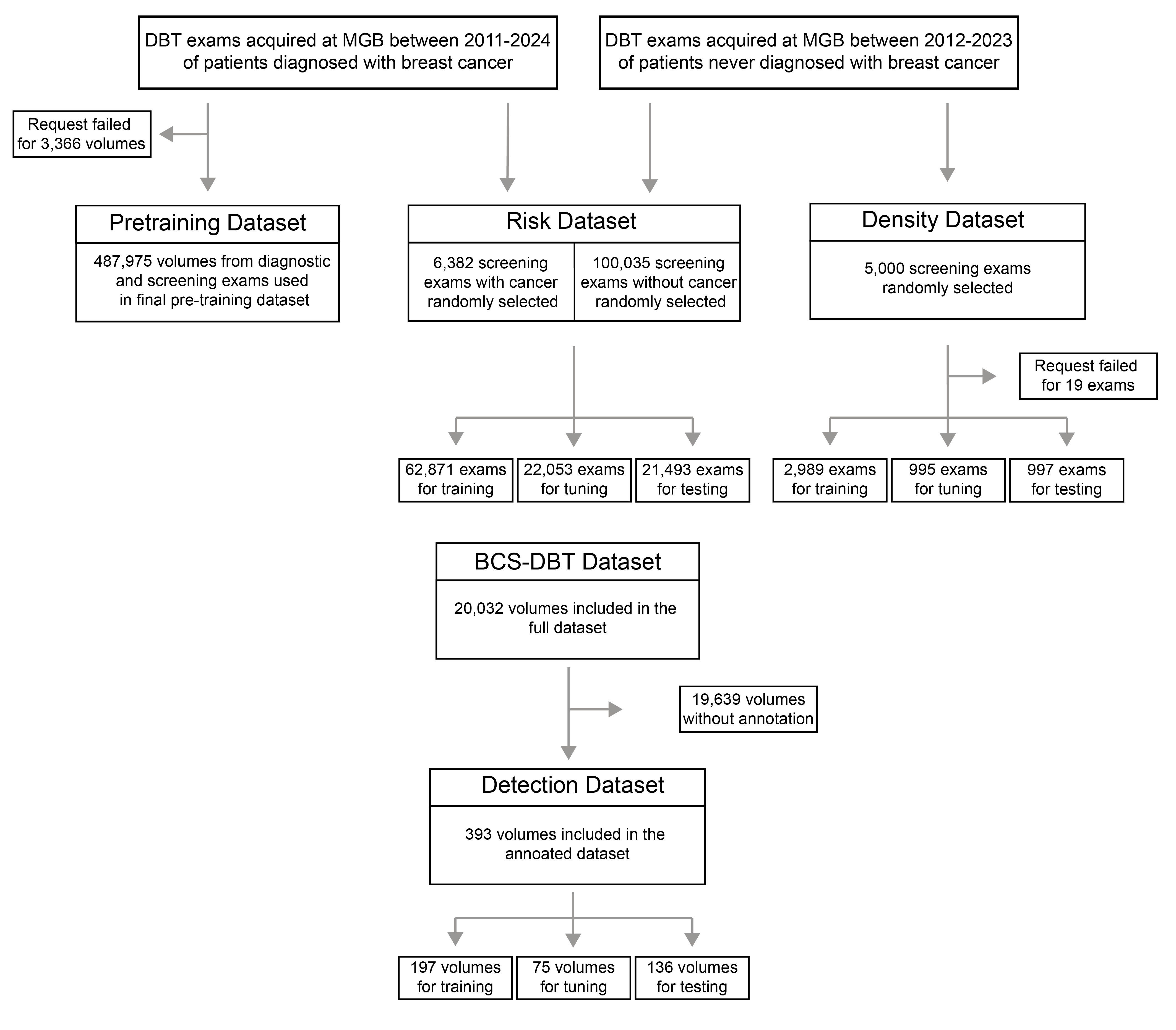}
    \caption{Patient Flow Chart for the individual datasets used in the study. `Risk Dataset' refers to the dataset used for predicting the risk of developing breast cancer. BCS-DBT: Breast cancer screening—digital breast tomosynthesis dataset. Throughout this work, an "exam" refers to a full imaging study (DICOM Study) comprising multiple 3D "volumes" (DICOM Series), also known as views.}
    \label{fig:patient_flowchart}
\end{figure}

\begin{figure}[ht]
    \centering
    \includegraphics[width=1\linewidth]{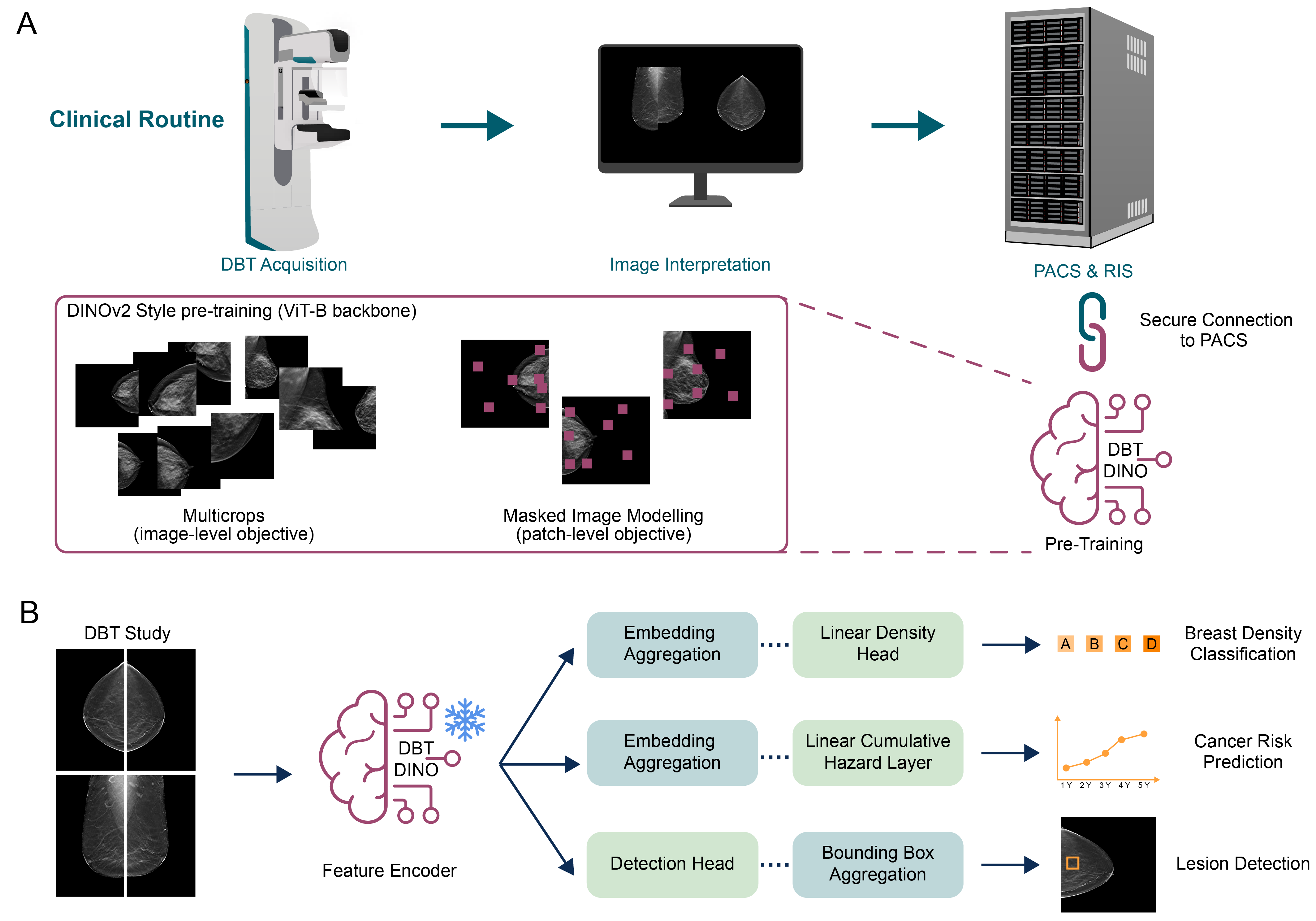}
    \caption{A: Overview of the pre-training process, through the PACS integration pre-training connects seamlessly into the clinical datastream. B: Overview of the downstream tasks.}
    \label{fig:enter-label}
\end{figure}
The institutional review board approved this Health Insurance Portability and Accountability Act (HIPAA)–compliant retrospective secondary analysis under IRB 2024P000452. The need for informed consent was 
waived.

\subsection{Cohort Description}
Four different datasets were used in this retrospective study. Three of the datasets were constructed from digital breast tomosynthesis (DBT) images acquired between 2011 and 2024 at the Mass General Brigham (MGB) hospital network in Boston, USA. 

The `pre-training dataset' included patients from MGB who carried the diagnosis of biopsy-proven breast cancer. Consequently, this dataset included healthy, pre-cancerous, and cancerous DBTs. A total of 3,227 patients were held out for testing, of which a subset with complete follow-up data was used for the risk prediction test set. All available DBT volumes from the remaining patients were included in pre-training.

The `density dataset' was comprised of "healthy" patients from MGB that have never received a breast cancer diagnosis. We randomly selected 5,000 unique screening exams from 5,000 patients containing all four standard views (bilateral CC and MLO) with equal representation across the four breast density categories, divided into 3,000 training, 1,000 validation, and 1,000 test exams.

The `risk dataset' included 31,561 patients.  All patients had at least 5 years of follow-up or a biopsy-proven diagnosis of breast cancer within 5 years. The dataset comprised 62,671 training, 22,053 validation, and 21,493 test exams, split at the patient level. There was no overlap between this task's "test" set and the "pre-training" dataset. 

The `detection dataset' was the Duke University (BCS-DBT) dataset containing DBT exams acquired between 2014 and 2018 with "region of interest" annotations for lesions suspicious for malignancy. \cite{buda_data_2021, buda_breast_2020}. Consistent with the DBTex challenge, we used the same splits as the challenge: 197 training, 75 validation, and 136 test volumes \cite{konz_competition_2023}. Data was downloaded from the Imaging Data Commons \cite{fedorov_nci_2021}.

Data splits and detailed statistics for all datasets are illustrated in Figure \ref{fig:patient_flowchart} and Table \ref{tab:pretraining_dataset_summary}, respectively.

\begin{table}[ht]
\centering
\caption{Dataset summary statistics for the four datasets used in the study. Information is provided as available for the particular dataset}
\label{tab:pretraining_dataset_summary}
\begin{tabular}{lcccc}
\hline
\textbf{Statistic} & \textbf{Pre-Training} & \textbf{Density} & \textbf{Risk} & \textbf{Detection}\\
\hline
Volumes                      & 487,975  & 19,924 & 425,668 & 393\\
Exams                     & 129,883 & 4,981  & 106,417 & 199\\ 
Patients                    & 27,990 & 4,981& 31,561 & 199\\ 
Age at study (mean $\pm$ SD)\,[years]       & 63.54 $\pm$ 11.64 & 57.76 $\pm$ 11.40 & 60.09 $\pm$ 10.47 & - \\
\hline
\textbf{Sex} & \\

Female                             & 26,735 (95.52\%) & 4855 (97.47\%) & 30,797 (97.57\%) & 199 (100\%)\\
Male                               & 83 (0.30\%)  & 1 (0.02\%) & 0  & 0\\
Unknown/Other                      & 1,172 (4.19\%)  & 125 (2.51\%) & 764 (2.42\%) &0\\
\hline
\textbf{Race} & \\
White                              & 22,954 (82.01\%) & 3772 (75.73\%) & 25,616 (81.17\%) & -\\
Asian                              & 1,176 (4.20\%) & 290 (5.82\%) & 1,245 (3.94\%) & -\\
Black                              & 1,106 (3.95\%) & 248 (4.98\%) & 1,934 (6.13\%) & -\\
Other/Unknown                      & 2,754 (9.84\%) & 671 (13.47\%) & 2,766 (8.76\%) & -\\
\hline
\end{tabular}
\end{table}

\subsection{Data Curation and Annotation}
Each patient's breast density was extracted from the corresponding radiology report using RegEx matching to identify the four BI-RADS breast density categories as defined in the 5th edition of BI-RADS \cite{sickles_acr_2013}.  
Each patient's breast cancer status and date of diagnosis was extracted from patient-matched pathology reports using a privacy-preserving open-source Large Language Model (LLM) implementation \cite{dorfner_comparing_2024}.

\subsection{Pre-Training}
DBT-DINO was initialized from the publicly released checkpoint of Meta AI's ImageNet-pretrained DINOv2, and continuously pre-trained using the DINOv2 algorithm \cite{oquab_dinov2_2024} with the "pre-training dataset". Hyperparameters including local crop size and scale were adjusted following published medical imaging adaptations of DINOv2 \cite{perez-garcia_exploring_2025}. To efficiently leverage such a large scale pre-training dataset, images were streamed directly from the clinical PACS system into the model training pipeline.

\subsection{Downstream Training} 
For all three downstream tasks, we used the pre-trained ViT backbone (e.g., DINOv2 vs. DBT-DINO) as a frozen feature encoder and trained only the task-specific prediction head.

The density and risk prediction tasks were both assessed at the study level and shared the same image encoding process. Each study contained the 2 standard views: craniocaudal (CC) and mediolateral oblique (MLO), for both breasts. Each slice was encoded using the DINO backbone, producing 1 CLS token and 1,369 patch tokens per 2D image. These tokens were aggregated across slices and views using per-feature mean and standard deviation. For density classification, this combined representation was passed to a linear layer predicting breast density class. For risk prediction, the prediction head was implemented as a cumulative hazard layer similar to \cite{yala_toward_2021}, outputting five risk probabilities for years 1-5 following the study date. For the density task, we additionally evaluated an ImageNet pre-trained DenseNet-121 model from torchvision as a CNN baseline with similar parameter count as DBT-DINO.

The detection task was performed at the volume level. Since the BCS-DBT dataset provides bounding boxes for only one slice per volume, only annotated slices were used during training. At inference, predictions were performed on all slices and aggregated to the volume level. The detection model consists of the frozen pre-trained ViT backbone, a neck creating a hierarchical feature pyramid from encoded image patches \cite{li_exploring_2022}, and a RetinaNet-style detection head \cite{lin_focal_2018}.

\subsection{Statistics}
95\% CIs for the performance metrics were estimated using bootstrapping with 1,000 repetitions. McNemar's test was used to compare performance of individual models on the detection and density tasks, and DeLong's test was used to compare AUC values for the risk prediction task. p < .05 was considered indicative of a statistically significant difference. Multiple testing was corrected using the Benjamini-Hochberg procedure.

\subsection{Packages used}
Training was performed in Python using PyTorch (V.1.13.1) with PyTorch Lightning (V.2.2.0), MONAI (V.1.3.0) and Highdicom (V.0.25.1) \cite{falcon_pytorch_2019, cardoso_monai_2022, bridge_highdicom_2022}. For data analysis and visualization NumPy (V.1.26.4), pandas (V.2.2.2), scikit-learn (V.1.5.0), matplotlib (V.3.9.0) and seaborn (V.0.13.2) were used \cite{harris_array_2020, mckinney_data_2010, pedregosa_scikit-learn_2011, hunter_matplotlib_2023, waskom_seaborn_2021}. Further implementation details are provided in supplementary materials. The code used in this study is available at: \href{https://github.com/QTIM-Lab/DBT-DINO/tree/main}{www.github.com/QTIM-Lab/DBT\-DINO}. 

\section{Results}
The pre-training dataset comprised 487,975 DBT volumes from 27,990 patients. The density dataset included 19,924 volumes from 4,981 patients, balanced across four BI-RADS density categories. The risk dataset contained 425,668 volumes from 106,417 exams across 31,561 patients, including 6,382 exams with subsequent cancer and 100,035 without cancer. The detection dataset consisted of 393 volumes from 199 patients from the Duke BCS-DBT dataset. Detailed demographics are provided in Table \ref{tab:pretraining_dataset_summary} and Figure \ref{fig:patient_flowchart}.

\subsection{Embedding Aggregation Strategies on the Density Task}
To make a per-volume prediction, the token embeddings for the individual 2D slices were aggregated. We evaluated four aggregation methods for the density prediction task as shown in \ref{tab:combining_embeddings}. We calculated aggregated embeddings (768 dimensions) for each view, concatenated them across views, and passed the resulting feature vector (3072 dimensions for 2 views per breast) to the linear layer for classification. To prevent overfitting for subsequent experiments we evaluated this experiment on the validation dataset, not on the held-out test dataset that was used for the other density experiments. Overall, we observed that adding the standard deviation of the embeddings improved performance. Mean and standard deviation of patch tokens achieved higher accuracy scores than CLS tokens, with 0.79 and 0.77 respectively, though differences between aggregation methods were not statistically significant. We used the mean and standard deviation of patch tokens as our aggregation strategy for all subsequent experiments.

\begin{table}[ht]
    \centering
    \caption{Performance for DBT-DINO on density classification with different embedding aggregation strategies. The aggregation strategies were: 1)  Mean of the CLS tokens for all slices; 2) Mean and standard deviation of the CLS tokens for all slices; 3) Mean of the patch tokens across all patches in the volume; 4) Mean and standard deviation of the patch tokens across all patches in the volume. Results are reported on the validation set.}
    \begin{tabular*}{\textwidth}{@{\extracolsep{\fill}}ccccc}\toprule
         &  CLS token Mean&  CLS token Mean + STD&  Patch Token Mean& Patch Token Mean + STD\\\midrule
         Validation Accuracy &  0.77 &  0.77 &  0.79 & \textbf{0.79}\\ 
         (95\% CI) & (0.74-0.79) & (0.74-0.79) & (0.76-0.81) & \textbf{(0.76-0.81)}\\ \bottomrule
    \end{tabular*}
    \label{tab:combining_embeddings}
\end{table}

\subsection{Fractionation Experiments for Breast Density Prediction}
\begin{figure}[ht]
    \centering
    \includegraphics[width=1\linewidth]{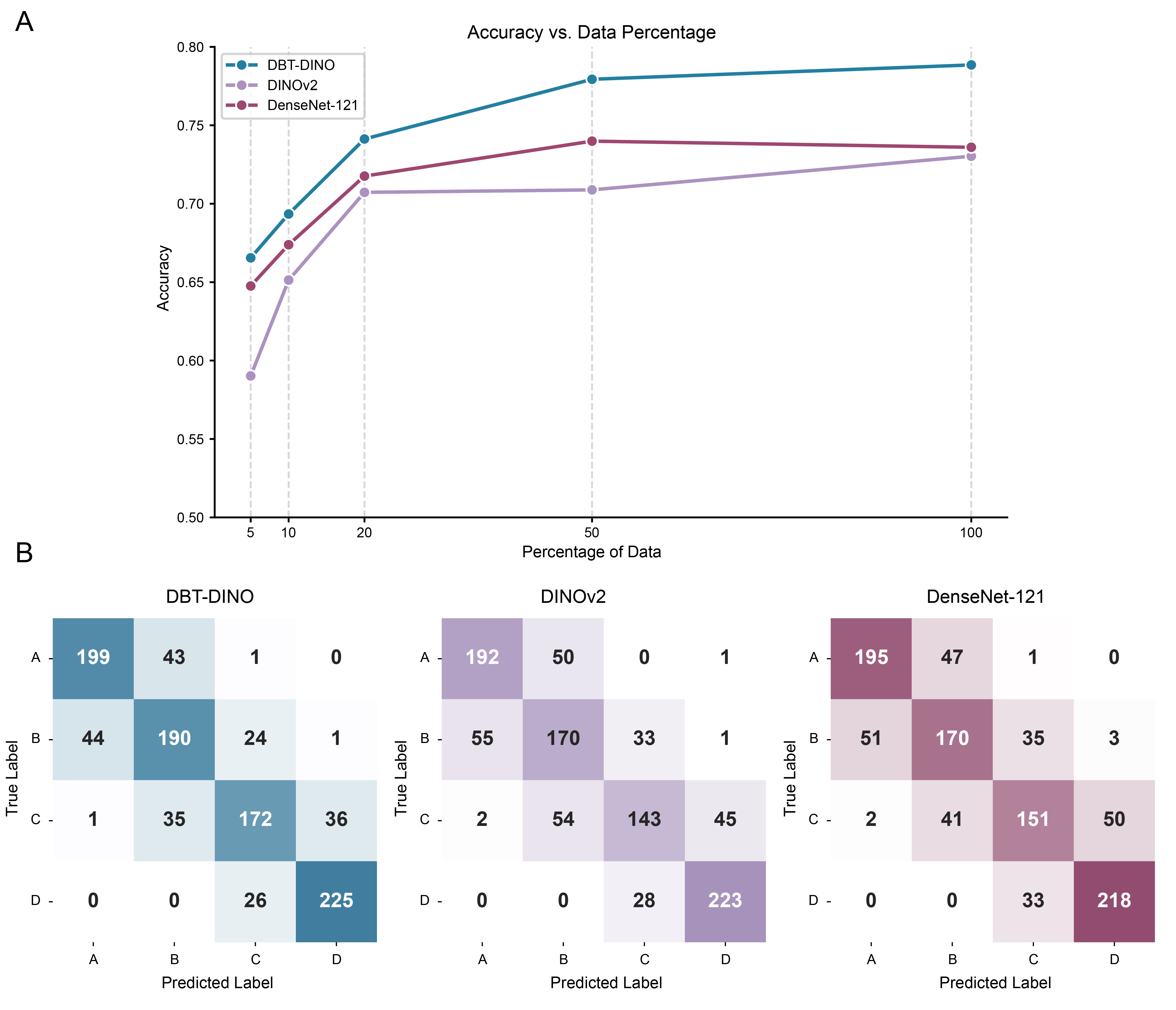}
    \caption{A: Fractionation on Breast Density 4-Class Classification. B: Confusion matrices for the DBT-DINO, DINOv2 and DenseNet-121 models trained on 100\% of the training data.}
    \label{fig:density_Fractionation}
\end{figure}

To evaluate whether the pre-training improved sample efficiency, we trained models on different fractions of the Density Dataset and evaluated their performance. The training fractions ranged from 5\% (150 samples; approximately 38 samples per class) to 100\% (2,989 samples).  Figure \ref{fig:density_Fractionation}a illustrates model performance of DBT-DINO, DINOv2 and the DenseNet-121 baseline on the held out test set of 997 exams. DBT-DINO consistently outperformed both DINOv2 and DenseNet-121 across most data fractions. The difference between DBT-DINO and DINOv2 was statistically significant at every fraction (all corrected p<.05). Compared to DenseNet-121, DBT-DINO had statistically superior performance at 50\% (corrected p=.02) and 100\% (corrected p<.001) of training data.
With the entire training set, DBT-DINO achieves its highest accuracy of 0.79 (95\% CI: 0.76-0.81), outperforming DenseNet-121 (0.74; 95\% CI: 0.71-0.76), and DINOv2 (0.73; 95\% CI: 0.70-0.76).

Misclassifications almost exclusively occur between adjacent classes---the reference standard label is `A' and the model predicts `B', or the label is `D' and the model predicts `C', as illustrated in Figure \ref{fig:density_Fractionation}b.

\subsection{Bias and Fairness Analysis on Breast Density Prediction}
Building upon prior studies that have reported differences in performance across racial subgroups, we evaluated potential bias using the breast density classification task and the model trained on 100\% of the training data. We restricted the analysis to healthy patients to avoid overlap with the pre-training dataset. Demographic information was available for 27,143 patients with 94,536 individual exams. Performance was evaluated across ethnicity and age subgroups. 

\begin{figure}[ht]
    \centering
    \includegraphics[width=1\linewidth]{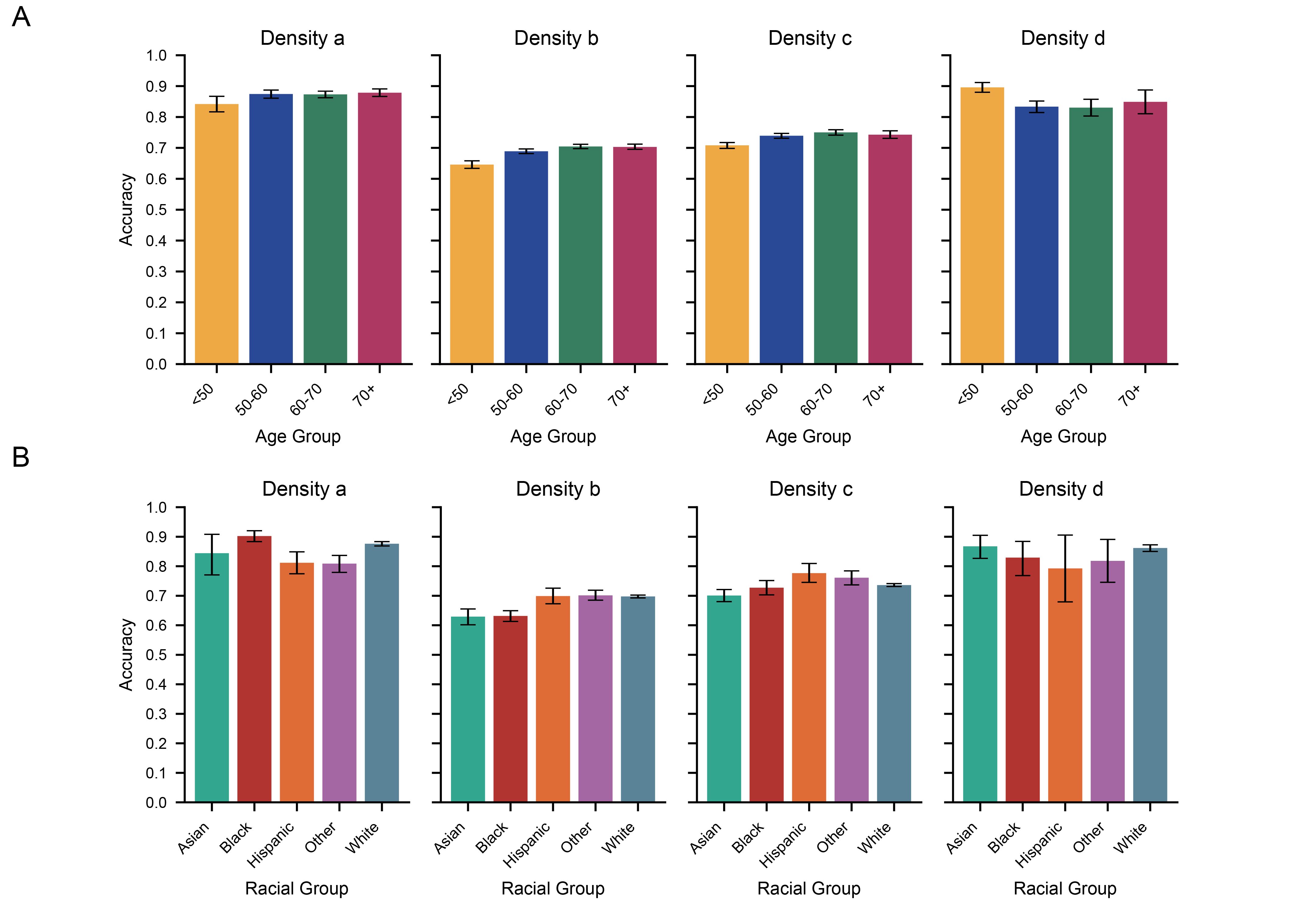}
    \caption{Accuracy of the DBT-DINO model on predicting breast density on a held-out test set. Performance groups based on Age and Ethnicity, stratified by their reference standard annotation of the 4 breast density categories: a: Entirely fatty tissue b: Scattered fibroglandular density areas c: Heterogeneously dense breasts d: Extremely dense breasts. \textbf{A}: Accuracy for different Age groups, \textbf{B}: Accuracy for different race and ethnic groups. The Other category contains patients with the following race entries in the electronic health record: 'Other', 'Unknown/Missing', 'Declined', 'Two or More', 'Native Hawaiian or Other Pacific Islander',
'American Indian or Alaska Native'}
    \label{fig:fairness}
\end{figure}

Overall, DBT-DINO did not show systematic performance bias for specific ethnic or age subgroups. Although accuracy differed between groups (0.70 to 0.74 for ethnic subgroups; 0.71 to 0.74 for age groups; full results in Supplemental Table \ref{tab:subgroup_density}), breast density distributions also varied substantially across these subgroups, consistent with prior literature \cite{kerlikowske_impact_2023}. Figure \ref{fig:fairness} shows the accuracy of the model for different subgroups stratified by the reference standard breast density label. It should be noted that 81\% of the patients in this dataset were white, which may temper conclusions. Figure \ref{fig:fairness} also shows that the model performs better on correctly classifying density for the extreme density categories (A and D) than for the intermediate categories (B and C). When predictions in the intermediate categories do not match the reference standard, they are almost always classified as adjacent categories on the density spectrum, as shown in Figure \ref{fig:density_Fractionation}B. For DBT-DINO only 3 (0.30\%) out of the 997 predictions are off by more than one density category. This pattern likely reflects inter-rater variability in reference standard annotations between radiologists \cite{portnow_persistent_2022}.

\subsection{Breast Cancer Risk Prediction}
DBT-DINO achieved a slightly higher average AUROC compared to DINOv2 on the breast cancer risk prediction task task with scores of 0.74 (95\% CI: 0.71–0.76) and 0.73 (95\% CI: 0.71–0.76), respectively. The year specific performance is also shown in \ref{tab:risk_year_performance}.

\begin{figure}[ht]
    \centering
    \includegraphics[width=1\linewidth]{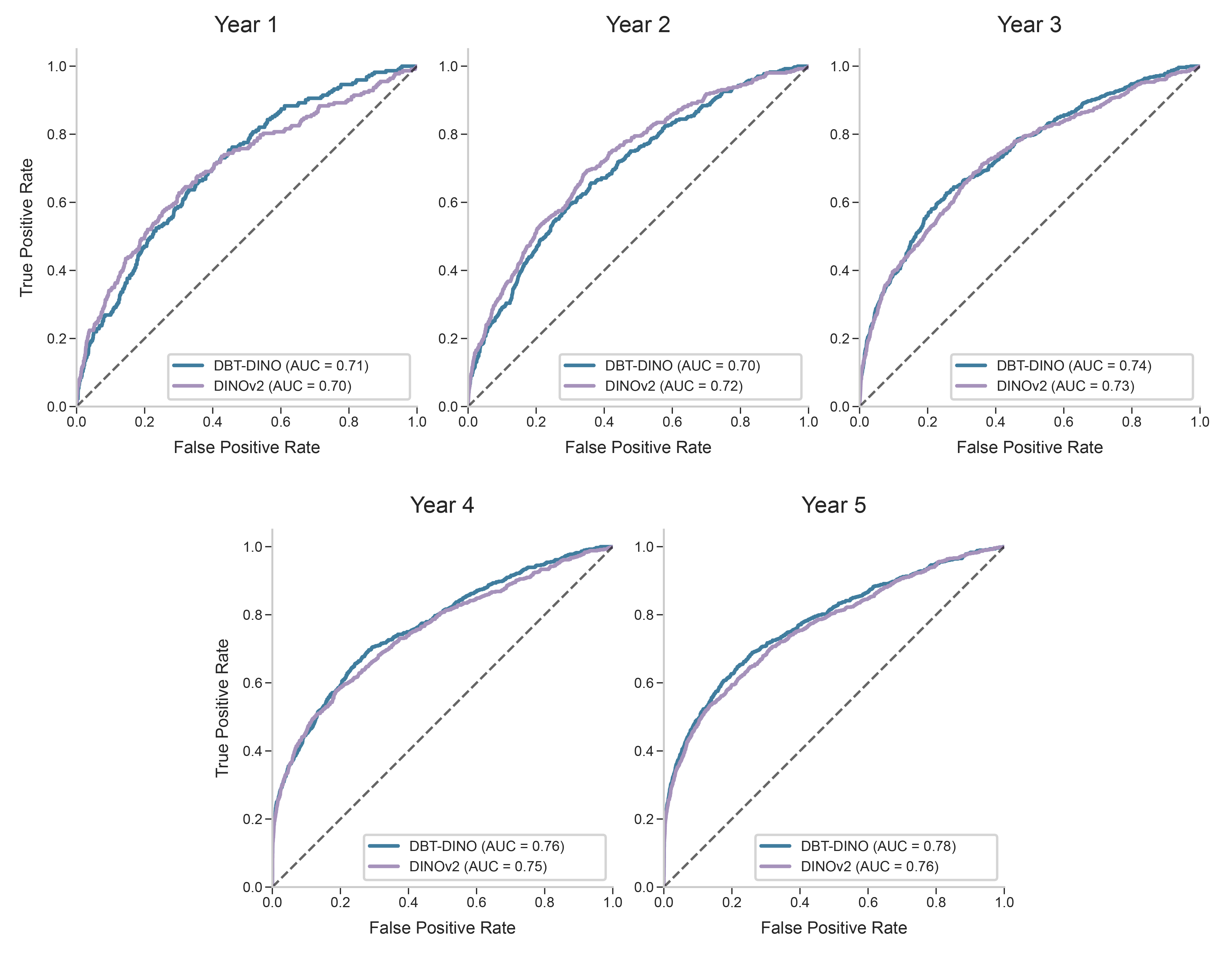}
    \caption{ROC Curves with AUROC scores for DBT-DINO and FB-DINO on the 5 year breast cancer risk prediction task test set.}
    \label{fig:risk}
\end{figure}

Figure \ref{fig:risk} shows ROC plots for both models and each year from 1 to 5. For both models, performance increased for later years of cancer occurrence, with both models achieving their highest AUROC at year 5: 0.78 AUROC for DBT-DINO (95\% CI: 0.76-0.80) and 0.76 (95\% CI: 0.74-0.78) AUROC for DINOv2. The performance difference between DBT-Dino and Dinov2 for 5 year risk was not considered statistically significant (p=0.057).

Table \ref{tab:subgroup_analysis} shows the subgroup analysis of the risk prediction performance for DBT-DINO stratified by breast density. Model performance decreased with increasing breast density (Table \ref{tab:subgroup_analysis}. Specifically, for Year 5, AUROC was 0.78 (95\% CI: 0.69--0.86) for BIRADS-A breast density, but decreased to 0.68 (95\% CI: 0.56--0.80)for BIRADS-D breast density for DBT-DINO. 

\begin{table}[ht]
\centering
\caption{Year-specific AUROC performance for breast cancer risk prediction on the test set.}
\label{tab:risk_year_performance}
\begin{tabular}{lccc}
\toprule
\textbf{Year} & \textbf{DBT-DINO AUROC (95\% CI)} & \textbf{DINOv2 AUROC (95\% CI)} \\
\midrule
1 & 0.71 (0.68--0.74) & 0.70 (0.66--0.74) \\
2 & 0.70 (0.67--0.72) & 0.72 (0.70--0.75) \\
3 & 0.74 (0.72--0.76) & 0.73 (0.71--0.75) \\
4 & 0.76 (0.74--0.79) & 0.75 (0.73--0.77) \\
5 & 0.78 (0.76--0.80) & 0.76 (0.74--0.78) \\
\midrule
Macro Avg. & 0.74 (0.73--0.75) & 0.73 (0.72--0.74) \\
\bottomrule
\end{tabular}
\vspace{0.3cm}
\end{table}

\begin{table}[ht]
\centering
\caption{Risk prediction performance based on breast-density subgroups for DBT-DINO on the test set. Percentages in brackets give the proportion relative to the corresponding total across all four density groups. Density labels: \textbf{a} – almost entirely fatty, \textbf{b} – scattered fibroglandular densities, \textbf{c} – heterogeneously dense, \textbf{d} – extremely dense.}
\begin{tabular}{lcccc}
\toprule
\textbf{Breast Density}          & \textbf{a} & \textbf{b} & \textbf{c} & \textbf{d} \\
\midrule
\textbf{\# Patients (total)}     & 835 (12\%)  & 3,258 (47\%) & 2,595 (37\%) & 513 (7\%) \\
\textbf{\# Pre-cancer Exams}  & 43  (6\%)   & 386 (49\%)   & 325 (42\%)   & 28 (4\%)  \\
\textbf{\# Healthy Exams}     & 2,108 (19\%)  & 9,936 (47\%) & 7,497 (35\%) & 1,170 (5\%) \\
\textbf{\# DBT Exams}          & 2,151       & 10,322       & 7,822        & 1,198     \\
\midrule
\textbf{Year 1 AUROC (95\% CI)}            & 0.71 (0.59--0.81) & 0.69 (0.64--0.74) & 0.72 (0.67--0.77) & 0.63 (0.31--0.89) \\
\textbf{Year 2 AUROC (95\% CI)}            & 0.63 (0.53--0.75) & 0.71 (0.68--0.75) & 0.69 (0.65--0.72) & 0.59 (0.40--0.77) \\
\textbf{Year 3 AUROC (95\% CI)}            & 0.79 (0.70--0.86) & 0.74 (0.71--0.77) & 0.72 (0.68--0.75) & 0.68 (0.51--0.82) \\
\textbf{Year 4 AUROC (95\% CI)}            & 0.78 (0.69--0.86) & 0.77 (0.74--0.80) & 0.75 (0.72--0.78) & 0.71 (0.58--0.83) \\
\textbf{Year 5 AUROC (95\% CI)}            & 0.78 (0.69--0.86) & 0.78 (0.76--0.81) & 0.76 (0.73--0.79) & 0.68 (0.56--0.80) \\
\midrule
\textbf{Macro Avg. AUROC}     & 0.74 (0.70--0.78) & 0.74 (0.72--0.76) & 0.73 (0.71--0.74) & 0.66 (0.56--0.74) \\
\bottomrule
\end{tabular}
\vspace{0.3cm}
\label{tab:subgroup_analysis}
\end{table}

\subsection{Lesion Detection}
Performance for the lesion detection task was assessed using sensitivity at different False Positive (FP) thresholds, as described in \cite{konz_competition_2023}. Average sensitivity reported was calculated across 1-4 FP per volume.  DINOv2 achieved an average sensitivity of 0.67 (95\% CI: 0.60--0.74), while DBT-DINO achieved an average sensitivity of 0.62 (95\% CI: 0.53--0.71). The Sensitivities ranged from 0.52 (95\% CI: 0.43--0.62) at 1 FP to 0.75 (95\% CI: 0.68--0.82) at 4 FP and 0.45 (95\% CI: 0.35--0.58) at 1 FP to 0.72 (95\% CI: 0.65--0.80) at 4 FP, respectively. 

\begin{figure}[ht]
    \centering
    \includegraphics[width=1\linewidth]{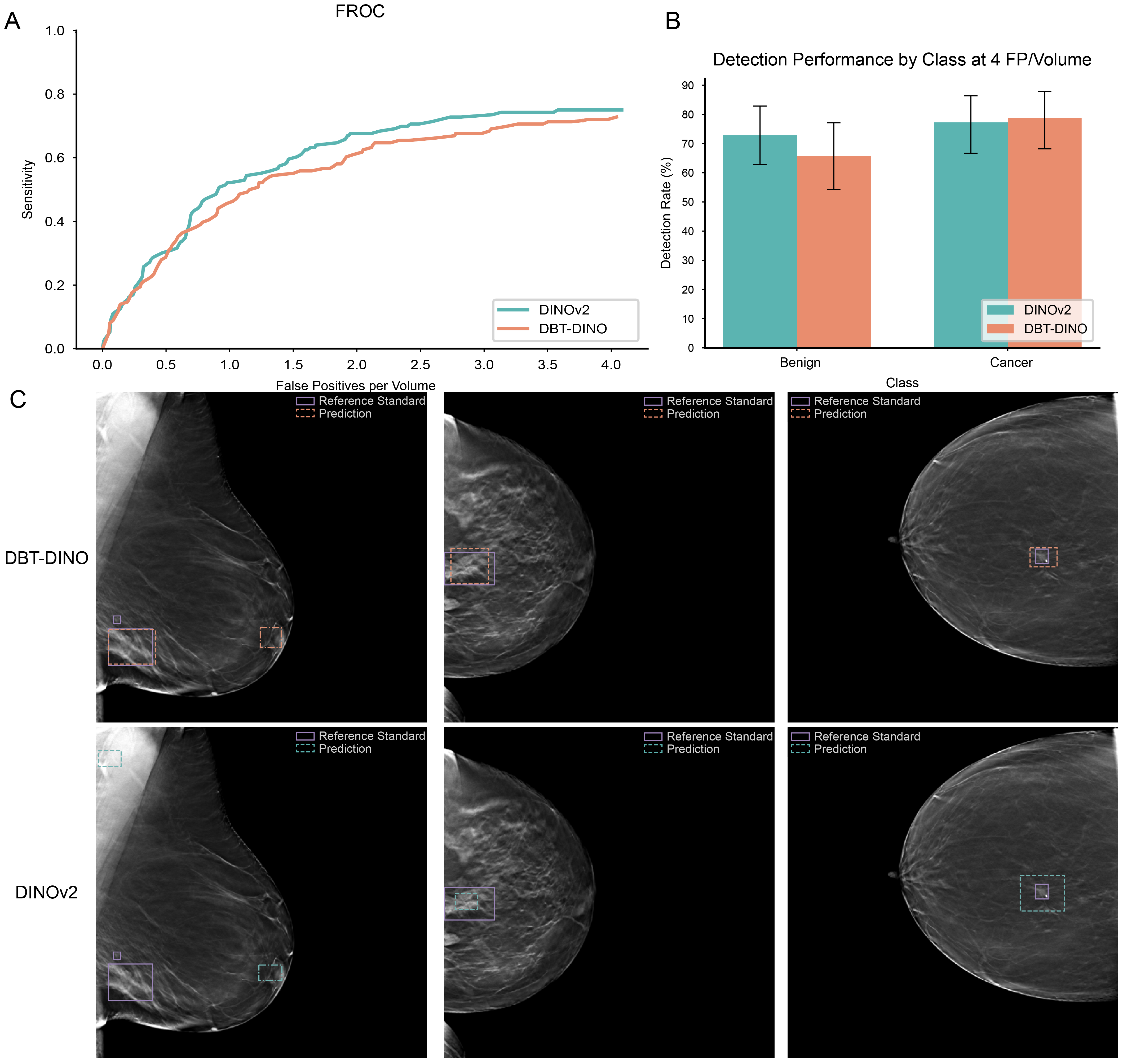}
    \caption{Free-response Receiver Operating Characteristic (FROC) curve for DBT-DINO and DINOv2 on the Detection Task}
    \label{fig:froc}
\end{figure}

Figure \ref{fig:froc}B illustrates the comparison between the detection performance for both models at 4 FP per volume. At this operating point, DBT-DINO detected 65.5\% of lesions that were ultimately found to be benign, compared to 72.9\% for DINOv2. For cancerous lesions, DBT-DINO achieved a detection rate of 78.8\% compared to 77.3\% for DINOv2. Figure \ref{fig:froc}C shows bounding box predictions from both models for three selected cases. The difference between models on the lesion detection task was not statistically significant (McNemar's test, p=0.60).

\section{Discussion}

The adoption of self-supervised pre-training with its ability to leverage large scale unlabeled data presents a paradigm shift for the development of AI models in medical imaging. However, it has remained largely unexplored for Digital Breast Tomosynthesis (DBT). To address this gap, we developed DBT-DINO, the first foundation model for DBT, a clinically important imaging modality with increasing adoption in breast cancer screening. Self-supervised pre-training was performed on 487,975 DBT volumes from 27,990 patients, and performance was evaluated across three clinically relevant tasks compared to ImageNet-pretrained baselines. For breast density classification, DBT-DINO achieved an accuracy of 0.79 (95\% CI: 0.76--0.81), significantly outperforming both DINOv2 (0.73, 95\% CI: 0.70--0.76; all corrected p<.05) and DenseNet-121 (0.74, 95\% CI: 0.71--0.76; corrected p<.001 at full training data). For 5-year breast cancer risk prediction, DBT-DINO achieved an AUROC of 0.78 (95\% CI: 0.76--0.80) compared to DINOv2's 0.76 (95\% CI: 0.74--0.78), though this difference was not statistically significant (p=0.574). For lesion detection, domain-specific pre-training showed mixed results: DINOv2 achieved higher average sensitivity at 0.67 (95\% CI: 0.60--0.74) compared to DBT-DINO with 0.62 (95\% CI: 0.53--0.71, p=.60), while DBT-DINO demonstrated superior performance specifically on cancerous lesions.

While no foundation models for DBT exist for comparison, individual sub-tasks have been previously tackled by single-task AI models. A study by Sexauer et al.\ trained a deep convolutional neural network (DCNN) on 3224 DBT images on the task of density prediction. They posed the problem as a binary task, classifying between BI-RADS A/B vs.\ C/D \cite{sexauer_diagnostic_2023}. They achieved an accuracy of 0.90 on their internal test dataset. After transforming our predictions into this binary format, DBT-DINO surpasses this performance with an accuracy of 0.94 using 100\% of the training data and can match their performance with an accuracy of 0.91 using only 10\% of our training data - 1/6th of the training samples Sexauer et al.\ used. While their model was fully fine-tuned on this task, we only trained a single linear layer, showcasing the expressiveness of our feature backbone. 

Yala et al.\ developed a deep learning model for breast cancer risk prediction on mammography using full model fine-tuning \cite{yala_toward_2021}. Our DBT-DINO model achieved comparable performance using only linear probing on frozen features. While direct comparison is limited by differences in imaging modality and dataset construction, this highlights the potential of DBT-DINO as a backbone for future studies on predicting breast cancer risk from DBTs.

For lesion detection, we can compare performance to the BCS-DBT challenge using the same dataset \cite{konz_competition_2023}. Both DBT-DINO (0.62) and DINOv2 (0.67) substantially outperformed the challenge baseline models (0.38 and 0.44 average sensitivity)\cite{konz_competition_2023}. However, challenge participants using fully fine-tuned object detection architectures (YOLOv5, Faster R-CNN) achieved superior performance with average sensitivities of 0.79-0.81. Since our approach applied only a detection head on frozen features rather than full fine-tuning of detection-specific architectures, this performance gap represents a limitation that highlights the need for future research on pre-training methods that can improve feature resolution and downstream performance on localized tasks, especially in medical imaging.

Several previous studies have developed biomedical foundation models based on DINOv2 and show benefits of pre-training \cite{chen_towards_2024, perez-garcia_exploring_2025}. These papers however only provide limited or no comparisons to the ImageNet baseline models. Furthermore, other work shows strong performance of the ImageNet DINOv2 model on a radiological task \cite{baharoon_evaluating_2024}. While we can achieve performance improvements on global tasks such as breast density prediction and risk prediction, performance on very localized tasks such as lesion detection did not benefit from our in-domain continued pre-training. 

This study has several limitations. First, the dataset used for the detection task was substantially constrained, with only 197 training volumes available for training from the Duke BCS-DBT dataset. This limited sample size may have substantially affected the model's performance on lesion detection, as deep learning approaches typically require larger annotated datasets to achieve optimal performance, particularly for complex tasks.
Second, to the best of our knowledge, besides the Duke BCS-DBT dataset used for lesion detection, there are currently no other publicly available DBT datasets that could be used to evaluate our model on external test data. We attempted to mitigate this limitation by employing large test datasets and conducting a bias and fairness evaluation across demographic subgroups. Furthermore, DBT-DINO is released as fully open-source and we encourage other researchers to use and evaluate our model.

Models trained using self-supervised learning demonstrate substantial potential for the analysis of Digital Breast Tomosynthesis. The in-domain pre-training using a state of the art self-supervised learning technique and the largest described dataset of DBT images showed variable effects across tasks, with benefits for some applications but potential limitations for others, particularly lesion detection tasks. Future research should explore new pre-training techniques specifically focused on medical imaging to enhance performance on tasks with highly localized information.

\FloatBarrier
\printbibliography
\section{Acknowledgments}
Much of the computation required for this research was performed on hardware generously provided by the Massachusetts Life Sciences Center (https://www.masslifesciences.com/).
This work has received support by the William G. Kaelin, Jr., M.D., Physician-Scientist Award of the Damon Runyon Cancer Research Foundation (PST-36-21), American Association for Cancer Research Breast Cancer Research Fellowship (21-40-49-KIM),  American Brain Tumor Association Basic Research Fellowship In Honor of Paul Fabbri, American Society of Clinical Oncology/Conquer Cancer Young Investigator Award, American Cancer Society Institutional Research Grant (IRG-21-130-10), a National Cancer Institute K12CA090354, and a National Institute of Biomedical Imaging and Bioengineering K08EB037077.

\section{Supplemental Material}

\begin{table}[ht]
\centering
\caption{Accuracy of DBT-DINO on the 4-class breast density classification task for different ethnic and age subgroups. The Other category contains patients with the following race entries in the electronic health record: 'Other', 'Unknown/Missing', 'Declined', 'Two or More', 'Native Hawaiian or Other Pacific Islander',
'American Indian or Alaska Native'. Age refers to age at image acquisition, so patients with scans across multiple years can appear in multiple age groups.}
\begin{tabular}{lccc}
\toprule
\textbf{Ethnic Subgroup} & \textbf{No. Exams} & \textbf{No. Patients} & \textbf{Accuracy (95\% CI)} \\
\midrule
Asian     & 3{,}495  & 1{,}058 & 0.695 (0.681--0.709) \\
Black     & 5{,}375  & 1{,}718 & 0.716 (0.704--0.727) \\
Hispanic  & 2{,}262  & 798     & 0.742 (0.724--0.760) \\
Other     & 4{,}880  & 1{,}626 & 0.735 (0.724--0.748) \\
White     & 78{,}524 & 21{,}943 & 0.736 (0.733--0.739) \\
\midrule
\textbf{Age Subgroup} & \textbf{No. Exams} & \textbf{No. Patients} & \textbf{Accuracy (95\% CI)} \\
\midrule
Age <50      & 17{,}417 & 7{,}449  & 0.711 (0.705--0.718) \\
Age 50--60   & 29{,}187 & 11{,}978 & 0.733 (0.728--0.738) \\
Age 60--70   & 29{,}009 & 11{,}235 & 0.742 (0.738--0.748) \\
Age 70+      & 19{,}179 & 7{,}000  & 0.740 (0.734--0.746) \\
\bottomrule
\end{tabular}
\vspace{0.3cm}
\label{tab:subgroup_density}
\end{table}

\subsection{Pre-Training}
For the pre-training the original implementation of DINOv2 was used with some adjustments to the hyperparameters. The following hyperparameters were used: Global Crop Scale 0.5-1.0, Local Crop Scale 0.2-0.5, Global Crop Size 518x518, Local Crop Size 196x196, Effective Batch Size 640. Training was run for 60{,}000 optimizer steps. The intermediate pre-training checkpoints were evaluated on the validation dataset of the density task to choose the best-performing checkpoint for subsequent experiments. The checkpoint used occurred about 29\% of the total training duration. Training was performed on four A100-40GB graphics cards (Nvidia, Santa Clara, USA) in using Pytorch's DistributedDataParallel (DDP). To efficiently leverage such a large scale pre-training dataset, images were accessed directly from the clinical PACS system via a custom reverse proxy server that authenticates user requests and ensures studies are within a pre-determined IRB protocol before forwarding the request over standard DICOM-web protocols. To make efficient use of this mechanism, we used a smaller, rotating local cache of downloaded files to ensure images were always available to feed to the GPUs.

\subsection{Downstream Training}
All downstream tasks utilize a frozen DBT-DINO ViT-B/14 backbone (patch size 14) for feature extraction from input images, with task-specific processing of the extracted features.

\subsubsection{Density Prediction}
We employed linear probing on frozen DINOv2 ViT-B/14 embeddings to classify breast density into four BI-RADS categories (a, b, c, d) from all four standard mammographic views.

Images were resized to $518 \times 518$ pixels and normalized to $[0,1]$ range after loading from DICOM format. The frozen ViT backbone processed images in chunks of 75 slices due to memory constraints. For each view (left craniocaudal [LCC], right craniocaudal [RCC], left mediolateral oblique [LMLO], right mediolateral oblique [RMLO]), we extracted patch tokens (excluding the classification token) from the final transformer layer. Statistical aggregation across spatial and slice dimensions yielded view-level representations by computing mean and standard deviation statistics, resulting in a 768-dimensional feature vector per statistic per view. The four view-level representations were concatenated to yield a final feature vector of dimensionality 3,072 per statistic (4 views $\times$ 768 dimensions), or 6,144 when using both mean and standard deviation (2 statistics $\times$ 4 views $\times$ 768 dimensions).

For comparison, we additionally extracted features from a pre-trained DenseNet-121 model downloaded from torchvision, which had been pre-trained on ImageNet data according to the torchvision training protocol. Individual slices in a DBT volume were processed through the DenseNet-121 model, with the embedding before the final fully-connected layer used as output, resulting in a 1,024-dimensional embedding per slice. Analogous to the DINOv2 features, these embeddings were aggregated across the slice dimension using mean and standard deviation statistics and then concatenated across the four views, resulting in a 4,096-dimensional feature vector per statistic (4 views $\times$ 1,024 dimensions), or 8,192 when using both mean and standard deviation (2 statistics $\times$ 4 views $\times$ 1,024 dimensions).

To accelerate training, embeddings were pre-computed and cached to disk before linear probing. The frozen DINOv2 model was run once on each DBT volume to extract and save patch token embeddings for all slices and views as separate numpy arrays (indexed by patient ID), eliminating redundant forward passes through the frozen backbone during training.

A single linear layer mapped the concatenated view-level features to the 4 breast density classes, with no hidden layers or non-linearities, following standard linear probing protocol. The backbone weights remained frozen throughout training, with only the linear classification layer being optimized.

Training employed a batch size of 64 for 75 epochs using the AdamW optimizer (weight decay: 0.0001) with cosine annealing learning rate scheduling and cross-entropy loss for multi-class classification. Hyperparameter optimization was performed using Optuna with 24 trials to tune the learning rate within $[1 \times 10^{-5}, 1 \times 10^{-2}]$. Training was performed with 16-bit mixed precision on a single RTX 8000 GPU (Nvidia, Santa Clara, USA). No data augmentation was applied, as embeddings were pre-computed from original images.

Model performance was evaluated using multi-class accuracy and macro-averaged F1-score. The best-performing model was selected based on minimum validation loss and evaluated on a held-out test set.

\subsubsection{Breast Cancer Risk Prediction}
We employed linear probing on pre-extracted DINOv2 (DBT-DINO) features to predict 5-year cumulative breast cancer risk from bilateral screening examinations.

Pre-computed DINOv2 patch token embeddings (dimension 3,072) were aggregated using mean and standard deviation statistics across all patch tokens per mammographic view. For bilateral screening examinations (RCC, LCC, RMLO, LMLO views), features from all views were concatenated, resulting in an input feature dimension of 6,144 (2 statistics $\times$ 3,072 features).

The prediction head consists of a single fully connected layer mapping input features to 5 yearly hazard predictions. These hazards are transformed into cumulative probabilities using a summation approach with an upper triangular mask operation, followed by a complementary log-log transformation ($1 - \exp(-x)$) to ensure outputs are bounded between 0 and 1, enforcing monotonically increasing cumulative risk over time.

Training employed a batch size of 64 for 100 epochs using the AdamW optimizer (weight decay: $1 \times 10^{-4}$) with cosine annealing learning rate scheduling. A masked binary cross-entropy (BCE) loss with logits was employed, with the masking mechanism accounting for censored follow-up data by only computing loss for observed time periods. The loss is normalized per sample across valid (non-masked) time points to handle variable follow-up durations. Optuna was used for automated learning rate tuning with 24 trials in the range $[1 \times 10^{-5}, 1 \times 10^{-3}]$, with median pruning applied to terminate unpromising trials early. Embeddings were extracted using three NVIDIA A100-40GB GPUs, with subsequent training performed on a single RTX 8000 GPU (Nvidia, Santa Clara, USA).

A balanced validation set was constructed by matching the number of cancer-positive and cancer-negative cases to provide unbiased performance estimates. Model checkpoints were saved based on both minimum validation loss and maximum average AUROC across all 5 years. Performance was assessed using year-specific AUROC for 1-year through 5-year predictions, as well as a weighted average AUROC across all time points using PyTorch's MultilabelAUROC metric.

\subsubsection{Lesion Detection}
We employed a RetinaNet-style object detection architecture with a frozen DINOv2 ViT-B/14 backbone trained end-to-end for lesion detection.

The detection model consisted of three main components: (1) a frozen DINOv2 ViT-B/14 backbone for feature extraction, (2) a Feature Pyramid Network (FPN) neck generating multi-scale features at 4 pyramid levels (P3: $74 \times 74$, P4: $37 \times 37$, P5: $18 \times 18$, P6: $9 \times 9$), and (3) a RetinaNet-style detection head with separate classification and bounding box regression towers. The FPN neck reduced the backbone's 768-dimensional features to 256 channels using $1 \times 1$ convolutions, then generated pyramid levels through bilinear upsampling (P3), max pooling (P5), and strided convolution (P6). Each detection tower consisted of 4 convolutional layers with GroupNorm (32 groups) and ReLU activation, followed by dropout regularization before the final prediction layers. We employed a multi-scale anchor strategy with 9 anchors per spatial location, generated from 3 aspect ratios (0.5, 1.0, 2.0) and 3 scales (1.0, 1.26, 1.587). Base anchor sizes were pyramid-level specific: 16 pixels for P3, 32 for P4, 64 for P5, and 128 for P6, corresponding to respective feature map strides of approximately 7, 14, 29, and 58 pixels.

Images were resized to $518 \times 518$ pixels and normalized to $[0,1]$ range. Training employed a batch size of 64 with 16-bit mixed precision for 50 epochs using the AdamW optimizer with cosine annealing learning rate scheduling. The model was optimized using a weighted combination of focal loss for classification and smooth L1 loss for bounding box regression. The focal loss employed tunable $\alpha$ and $\gamma$ parameters to address severe class imbalance, with additional label smoothing for regularization. Positive anchors were assigned to ground truth boxes using IoU matching.

Hyperparameter optimization was performed using Optuna with 200 trials, tuning: learning rate ($1 \times 10^{-5}$ to $1 \times 10^{-2}$), weight decay ($1 \times 10^{-6}$ to $1 \times 10^{-3}$), label smoothing (0.0 to 0.1), dropout rate (0.0, 0.05, 0.1, 0.15), classification-to-bbox loss ratio (0.1 to 10.0), focal loss parameters ($\alpha$: 0.25-0.95, $\gamma$: 0.5-2.0), anchor assignment IoU thresholds (positive: 0.4-0.6, negative: 0.2-0.5), negative-to-positive anchor ratio (1-5), and non-maximum suppression threshold (0.03-0.3). Early stopping with patience of 10 epochs was applied, monitoring mean sensitivity at 1-5 false positives per image on the validation set.

Training augmentation included: random horizontal flip ($p=0.5$), random vertical flip ($p=0.5$), random zoom (factor: 0.8-1.5, $p=0.5$) with constraints ensuring post-zoom bounding box sizes remained within [15-206 pixels width, 9-182 pixels height], bbox-aware coarse dropout (15-20 rectangular regions of $20 \times 20$ to $40 \times 40$ pixels, $p=0.3$) that avoided overlapping with lesion locations, random contrast adjustment ($\gamma$: 0.8-1.2, $p=0.3$), and random Gaussian noise ($\sigma=0.05$, $p=0.3$).

Model performance was evaluated using mean sensitivity at 1, 2, 3, 4, and 5 false positives per image, following the challenge evaluation protocol. A predicted bounding box was counted as a true positive if its center fell within $\max(\frac{1}{2} \times \text{diagonal length of ground truth box}, 20 \text{ pixels})$ of a ground truth lesion center, with each ground truth lesion matched to at most one prediction. The model checkpoint with highest validation mean sensitivity was selected for final evaluation. 
\end{document}